\documentclass[conference]{IEEEtran}
\IEEEoverridecommandlockouts
\usepackage{cite}
\usepackage{amsmath,amssymb,amsfonts}
\usepackage{algorithmic}
\usepackage{graphicx}
\usepackage{textcomp}
\usepackage{xcolor}
\usepackage{multirow}
\usepackage{mathtools}
\usepackage{footnote}
\def\BibTeX{{\rm B\kern-.05em{\sc i\kern-.025em b}\kern-.08em
    T\kern-.1667em\lower.7ex\hbox{E}\kern-.125emX}}
\begin{document}

\title{Meta-DRN: Meta-Learning for 1-Shot Image Segmentation\\
% % \thanks{Identify applicable funding agency here. If none, delete this.}
}

\author{\IEEEauthorblockN{Atmadeep Banerjee}
\IEEEauthorblockA{\textit{Dept. of Computer Science} \\
\textit{Birla Institute of Technology and Science, Pilani}\\
Pilani, India \\
%atmadeepb@gmail.com}
f20170101@pilani.bits-pilani.ac.in}
% \and
% \and
}

\maketitle

\begin{abstract}
Modern deep learning models have revolutionized the field of computer vision. But, a significant drawback of most of these models is that they require a large number of labelled examples to generalize properly. Recent developments in few-shot learning aim to alleviate this requirement. In this paper, we propose a novel lightweight CNN architecture for 1-shot image segmentation. The proposed model is created by taking inspiration from well-performing architectures for semantic segmentation and adapting it to the 1-shot domain. We train our model using 4 meta-learning algorithms that have worked well for image classification and compare the results. For the chosen dataset, our proposed model has a 70\% lower parameter count than the benchmark, while having better or comparable mean IoU scores using all 4 of the meta-learning algorithms. 
\end{abstract}

% \begin{IEEEkeywords}
% Few-shot learning, Image segmentation, Meta-learning
% \end{IEEEkeywords}

\section{Introduction}
Modern deep learning models have made tremendous progress in the field of computer vision, even outperforming humans in some cases. However, a significant drawback of most of these models is their dependency on large amounts of labelled data. Recent progress in semi-supervised and few-shot learning algorithms is helping to alleviate this. The simplest idea that can be used in a situation where training data is scarce, is transfer learning\cite{transfer}. In several computer vision tasks, fine-tuning models pretrained on the ImageNet\cite{ImageNet} dataset has been shown to perform well, even with a small dataset size. However, in many situations, especially where the dataset to be fine-tuned on differs significantly from the pretraining dataset, transfer learning does not work too well\cite{transfusion}. The second approach is semi-supervised learning which can be used in situations where plenty of unlabelled data is available but labelled data is scarce. But in situations where extra unlabelled data is not available either, few-shot learning is the viable option.

The problem of few-shot learning can be formulated as a meta-learning problem. In such a scenario, a meta-learning model is trained on discrete tasks where, each task forms a data point for the meta-model. A task is defined as a combination of a support set $S$ and query set $Q$. $S$ has a small number of support examples $Is$ along with their labels $Ls$. $Q$ is a set of query examples $Iq$ and their labels $Lq$. The model has to predict $Lq$ given $<Is, Ls, Iq>$. 1-shot learning is an extreme case of few-shot learning where the support set $S$ is restricted to have a single training sample.

In practice, 1-shot learning is meaningful to use for testing the viability of a deep learning an approach without incurring the large cost of acquiring a labelled dataset. By extension, it is also meaningful for 1-shot models to be lightweight in terms of memory and computational power in order to minimize cost. For this reason we propose a model for 1-shot segmentation that is 70\% smaller than the dataset benchmark in terms of parameter count, and can be trained in a shorter time with a lower memory requirement.

The objective of this paper is two-fold. First we propose our novel lightweight model architecture for 1-shot image segmentation. Next we compare the results upon training using 4 gradient-based meta-learning algorithms that have shown good results in image classification. The chosen algorithms are MAML\cite{maml}, Meta-SGD\cite{msgd}, FOMAML\cite{maml} and Reptile\cite{reptile}. We use the FSS-1000 dataset\cite{fss} for training. We made the choice of using gradient-based meta-learning approaches instead of a metric-based approach as used in the FSS-1000 benchmark because it allowed us to train lightweight models, while retaining comparable accuracy. The chosen algorithms all belong to the MAML\cite{maml} family of algorithms. These algorithms use a \textit{meta model} to initialize a learner with a set of weights such that, given a task, the learner can converge in a minimal amount of training iterations. Our network architecture is designed to be lightweight. To ensure this, we use a model with minimal downsampling of input feature maps. We use dilated convolutions, instead of downsampling to increase the receptive field of our convolution layers. This approach allows us our model to be much smaller than an encoder-decoder architecture, while still being comparably performant. 

\section{Related Work}
% \begin{figure*}
%   \centering \includegraphics[width=0.8\textwidth]{MetaDRN.png}
%   \caption{The proposed model architecture.}
% \end{figure*}

Image segmentation is an extensively studied field in computer vision. The task requires dense pixel-wise predictions, given an input image. Encoder-decoder architectures are mostly commonly used for this task. The U-Net architecture, designed by Ronneberger et al.\cite{Unet} is a very popular architecture for image segmentation. It improves over standard encoder-decoder architectures by incorporating skip connections between the encoder and decoder, allowing deep models to be trained. Standard CNN models like Resnet\cite{Resnets} are often used as encoders in U-Net based architectures. These CNN models perform successive downsampling of feature maps to discard irrelevant information. This also has the effect of increasing the effective receptive field of the convolution kernels in deeper layers. While this is an effective approach for image classification, for dense pixel wise predictions, this discarding of information is harmful, especially when the object of interest is not spatially dominant.

Yu and Koltun suggest in \cite{msa} that standard the task of dense predictions is structurally different from image classification. They show that dilated convolutions perform better than standard CNNs on dense prediction tasks. In \cite{drn} Yu et al. introduce Dilated Residual Networks, which are a modification of standard ResNets\cite{Resnets} using dilated convolutions instead of downsampling of feature maps. This architecture achieved much better performance on tasks like image segmentation and object localization. We use the idea proposed in this paper and adapt it to the few shot setting. 

% Few-shot seg background and citations. Most existing work uses rel nets. These are heavier.
Most existing work in few-shot segmentation uses variants of branched CNN architectures\cite{old1,old2,old3,old4} or relational networks\cite{rel_net_2}. In general these methods rely on using learnable modules to find the similarity between support and query images. Although meta-learning has seen a lot of success with few-shot image classification, frequently reaching state-of-the-art results, these techniques have not been as extensively used for segmentation. Some examples include \cite{meta_old,meta_old2}. We use the FSS-1000\cite{fss} dataset for our models. This dataset is designed specifically for few-shot segmentation. Most previous works made use of the PASCAL-5\textsuperscript{i} dataset\cite{oslss}, which is a few-shot version of the PASCAL-VOC dataset\cite{pascal}. PASCAL-5\textsuperscript{i} only has 20 classes, which is far lower than FSS-1000. Li et al. argue in \cite{fss} that approaches trained on only 20 classes may suffer from strong bias. A drawback of this dataset is that most images contain a single object. Because of this, we have specifically visualized some test images that contain multiple objects in Figure 4.

\section{Studied Algorithms}
\textbf{MAML:} MAML trains a meta-learner to learn how to initialize a learner with a suitable set of initial weights $\theta$. When the learner is presented with the support set of a new task, it can converge in a small amount of gradient-descent iterations.
\begin{equation}
    \theta^{\prime} = \theta - \alpha . \nabla \mathcal{L}_{train(\mathcal{T})}(\theta) 
\end{equation}

Here $\alpha$ is a predefined constant. The meta-learner is trained to improve its initialization of learner weights using the query set(or test set) of training tasks.
\begin{equation}
    \theta = \theta - \beta . \nabla \mathcal{L}_{test(\mathcal{T})}(\theta^{\prime}) 
\end{equation}

Here $\beta$ is the scalar learning rate for the meta-learner. The meta-learner can be trained using any gradient based optimization algorithm like Adam.

\textbf{FOMAML:} As is evident from equation (2) MAML requires calculating a second-order derivative when optimizing the meta-learner weights, which leads to a significant increase in training time. Finn et al. show that even without calculating second-order derivatives, performance comparable to MAML can be reached for image classification and reinforcement learning. This algorithm is called First Order approximation to MAML or FOMAML. In FOMAML, equation (2) is modified by assuming that the gradient term in the calculation of $\theta^{\prime}$ is independent of $\theta$.

\textbf{Meta-SGD:} The third algorithm we train our model on is Meta-SGD\cite{msgd}. This algorithm is same as MAML except for the fact that the learning rates $\alpha$ for each parameter in the learner is not a constant. Instead it is a trainable parameter that is also optimized by the meta-learner. The meta-learner therefore has control over not only the initialization, but also the learning rates of the parameters of the learner. In practice, this means that Meta-SGD shows faster convergence, higher accuracy and lower sensitivity to hyperparameters than MAML. The learner's training equation is same as that of MAML, except for $\alpha$ not being a constant. The meta-learner is trained according to the following modified equation
\begin{equation}
    (\theta,\alpha) = (\theta,\alpha) - \beta . \nabla_{(\theta,\alpha)} \mathcal{L}_{test(\mathcal{T})}(\theta^{\prime}) 
\end{equation}

Like MAML, Meta-SGD also requires calculation of second-order derivatives and therefore has a higher training time and memory requirement.
 
\textbf{Reptile:} The final algorithm we use to train our model is Reptile\cite{reptile}. Reptile is similar to FOMAML in that it does not require calculation of second-order derivatives. The Reptile meta-learner also learns to initialize the weights $\theta$ of the learner. The learner performs $k (>1)$ gradient updates on the support set of a task.
\begin{equation}
    \begin{multlined}
    \textrm{repeat $k$ times:}\\
    \theta^{\prime} = \theta - \alpha . \nabla \mathcal{L}_{train(\mathcal{T})}(\theta) 
    \end{multlined}
\end{equation}

When training the meta-learner, Reptile does not need the training tasks to have a query set. It uses the difference between the final weights of the learner $\theta^{\prime}$ and the initial weights $\theta$ to train the meta-learner according to the following equation.
\begin{equation}
    \theta = \theta + \beta . (\theta^{\prime} - \theta)
\end{equation}

Nichol et al. also suggest that the term $\theta^{\prime} - \theta$ can be treated as the gradient for the meta-learner and it can be trained using any gradient-descent algorithm like Adam.

\begin{figure}[t]
\begin{center}
   \includegraphics[width=1.0\linewidth]{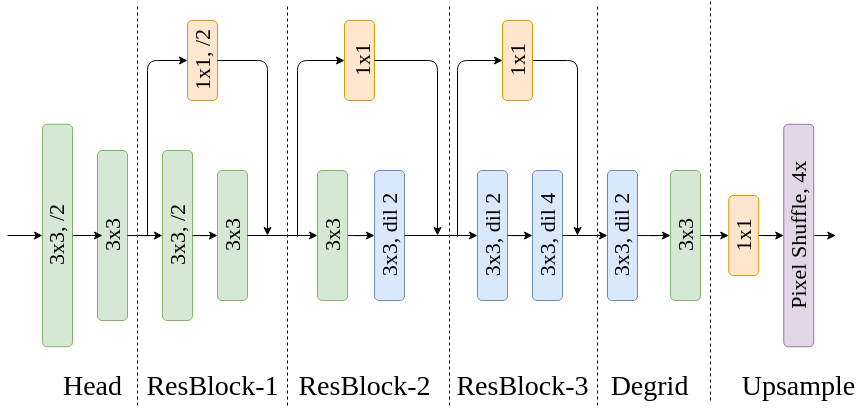}
\end{center}
   \caption{The proposed model architecture.}
\label{fig:long}
\label{fig:onecol}
\end{figure}

\section{Proposed Architecture}
% \subsection{Architecture}

\begin{figure*}
  \centering \includegraphics[width=1\textwidth]{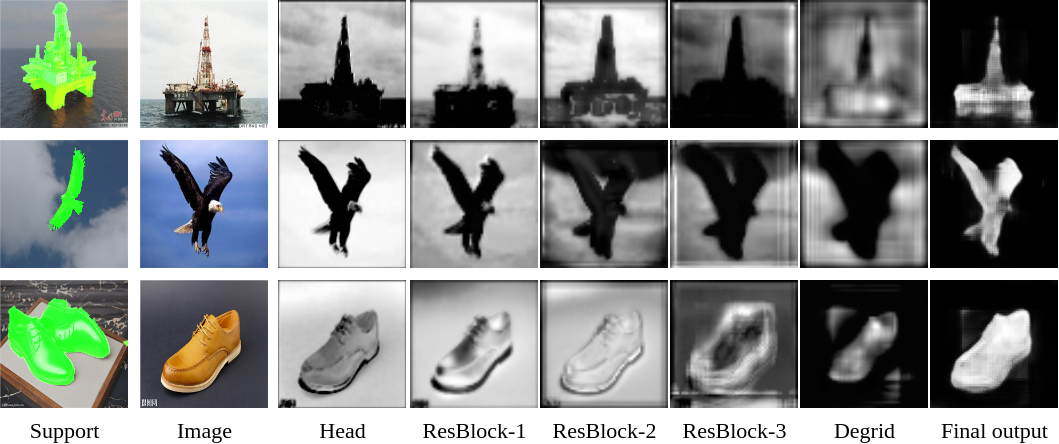}
  \caption{The activations of a MAML Meta-DRN on 3 test tasks. For each layer the feature map with the highest mean activation value is visualized }
\end{figure*}

 As already discussed, successive downsampling of feature maps is not a good choice for tasks requiring dense predictions and the alternative of using dilated convolutions to increase the receptive field is much more suitable. Also, since U-Net based models perform a learned upsampling of feature maps using deconvolutions, a network that performs less reduction of the feature map's size in the encoder, will have to use lesser parameters for upsampling in the decoder. A network with dilated convolutions is therefore capable of being more lightweight than a U-Net based model.

Our proposed architecture is depicted in Figure 1. It consists of a head, followed by 3 residual blocks (with skip connections), two degridding layers\cite{drn} and a final upsampling layer. The residual blocks are similar to those in ResNet-18 and ResNet-34, but has fewer layers and makes use of dilated convolutions. The upsampling layer makes use of a sub-pixel convolution. The feature map with highest activation, formed after passing through each layer group is visualized in Figure 2.

\textbf{Head: }The head consists of two 3x3 convolution layers, with each layer being followed by a batch normalization layer\cite{Batchnorm} and a leaky relu\cite{lrelu} non linearity. The first layer has a stride of 2 and 16 filters. The second layer has 64 filters.

\textbf{ResBlock-1:} The head is followed by ResBlock-1. It is similar to a normal ResNet residual block, having a stride 2 convolution followed by a normal 1-dilated convolution. All convolution layers have 128 filters.

\textbf{ResBlock-2:} In ResBlock-2, there is a 1-dilated convolution, followed by a 2-dilated convolution. All convolution layers have 256 filters. If the first convolution layer had performed a strided convolution, the effective receptive field of the second layer would have been doubled. Instead of performing a strided convolution and reducing the feature map size, we double the dilation of the second convolution layer to have the same effect. We can represent this by:

\begin{equation}
    (\mathcal{G}^{2}_{2} *_{2} f^{2}_{2})(\mathbf{p}) = \sum_{\mathbf{a+2b=p}} \mathcal{G}^{2}_{2}(\mathbf{a})f^{2}_{2}(\mathbf{b})
\end{equation}

Similar to the notation used in \cite{drn}, here $\mathcal{G}^{l}_{i}$ indicates the $i$\textsuperscript{th} filter of the $l$\textsuperscript{th} ResBlock. $f^{l}_{i}$ indicates the filter associated with layer $\mathcal{G}^{l}_{i}$. 

\textbf{ResBlock-3:} In ResBlock-3, there is a 2-dilated convolution followed by a 4-dilated convolution. All convolution layers have 512 filters. The first convolution layer of the this block would have double its receptive field had we performed a strided convolution in ResBlock-2. To have the same effect, this layer also has a 2-dilated convolution. Similarly, the second layer of the ResBlock-3, would have had 4 times its receptive field if there were strided convolutions in both ResBlock-2 and ResBlock-3. To have this same effect, we make the convolution 4-dilated.

\begin{equation}
    (\mathcal{G}^{3}_{1} *_{2} f^{3}_{1})(\mathbf{p}) = \sum_{\mathbf{a+2b=p}} \mathcal{G}^{3}_{1}(\mathbf{a})f^{3}_{1}(\mathbf{b})
\end{equation}

\begin{equation}
    (\mathcal{G}^{3}_{2} *_{4} f^{3}_{2})(\mathbf{p}) = \sum_{\mathbf{a+4b=p}} \mathcal{G}^{3}_{2}(\mathbf{a})f^{3}_{2}(\mathbf{b})
\end{equation}

\textbf{Degrid:} As observed by Yu et al.\cite{drn}, using dilated convolutions have a tendency to introduce grid artifacts in the feature maps. These are removed by using a degridding layers consisting of convolution layers with successively smaller dilation. The degridding layers in our architecture consists of two successive 3x3 convolutions with 2-dilation followed by 1-dilation. Both have 512 filters.

\begin{table*}[t]
\centering
\caption{Results}
\begin{tabular}{|c|c|c|c|c|}
\hline
Setting             & mIoU (thresh = 0.5)  & mIoU (thresh = 0.35)  & \begin{tabular}[c]{@{}c@{}}Time per epoch\\ (min)\end{tabular} & Parameter Count        \\ \hline
FSS-1000            & 73.47                & -                     & 48:00*                                                         & 32.83M                 \\ \hline
% EfficientLab        & 76.45 \pm 1.16       & -                     & -                                                              & -                      \\ \hline
% DoG-LSTM            & \textbf{80.83}                & -                     & -                                                              & -                      \\ \hline
Meta-DRN, MAML      & $75.17 \pm 0.12$       & $75.19 \pm 0.11$        & 28:40                                                          & \multirow{4}{*}{9.56M} \\ \cline{1-4}
Meta-DRN, FOMAML    & $68.21 \pm 0.18$       & $68.14 \pm 0.01$        & 18:09                                                          &                        \\ \cline{1-4}
\textbf{Meta-DRN, Meta-SGD}       & $75.89 \pm 0.39$       & \textbf{76.63} $\pm$ \textbf{0.32}        & 29:32                                   &                        \\ \cline{1-4}
Meta-DRN, Reptile   & $64.29 \pm 0.22$       & $63.98 \pm 0.18$        & 51:09                                                          &                        \\ \hline
\end{tabular}
\end{table*}

\textbf{Upsample:} A deviation from the original DRN model in our architecture, apart from model depth, is the use of a sub-pixel convolution layer\cite{PixelShuffle} for the upsampling of the final feature maps to create the ouput. The original DRN model uses either of bilinear upsampling or a deconvolution layer. In our model, sub-pixel convolution was chosen because it is faster for upsampling than standard deconvolution, while still having the advantage of learned upsampling. It performs a convolution in the low resolution space, and upsamples the result with a phase shift operation. Also, a sub-pixel convolution operation is mathematically equivalent to a deconvolution while still being faster.

% \subsection{Loss}
% The model is trained using cross entropy loss. 

% \begin{equation}
%     L = -\frac{1}{n}\sum_{i=1}^{n}(y_{i}\log(\hat{y}_{i}) + (1-y_{i})\log(1-\hat{y}_{i})
% \end{equation}

% A combination of cross entropy and dice loss \cite{dice} was also tried since it is known to perform well for segmentation tasks. Models trained using this combined loss initially improved faster on the validation metric, but were eventually surpassed by models trained with only cross-entropy loss.

\section{Experiments}
\subsection{Dataset}
The FSS-1000 dataset\cite{fss} consists of images belonging 1000 classes, with 10 images per class. The dataset has 10 labelled images per class, with dense pixel-wise labels. Out of the 1000 classes 240 has been marked by the authors as the test set. Out of the 760 train classes, 700 are used for training and 60 for validation. Final results are reported on the 240 test classes.

\subsection{Training}
All our models are trained in 1-shot setting for 200 epochs. Basic augmentations, including flip, rotate, zoom, warp and lighting transforms were applied to the images in the 700 training classes. Cross entropy loss was used to train all the models. The evaluation metric is mean Intersection over Union(IoU). For predicted and ground-truth masks, IoU becomes the area of intersection divided by the sum of areas of both masks reduced by area of union. 
\begin{equation}
    IoU = \frac{Pred * Targ}{Pred + Targ - Pred*Targ}
\end{equation}

All training runs are performed on an RTX 2080 GPU and the time per epoch on this GPU is reported for comparison. Note that the time per epoch for FSS-1000 model is based on our implementation. The training details for each algorithm are as follows: 

For MAML, the meta batch size was set to 5. The learning rate for the learner(fast weights) was initialized to 1e-3. The number of train steps for the learner(fast weights) were set to 1. The learning rate for the meta-learner was initialized to 1e-3 and set to halve every time the IoU plateaued for more than 8 epochs. Cosine annealing and One Cycle learning rate schedules were also tried but found to not significantly affect performance. AdamW\cite{AdamW} optimizer was used for training the meta-learner. We found that the weight decay introduced by AdamW to be beneficial. These setting were same for FOMAML and Meta-SGD.

For Reptile, the meta batch size was set to 8. The learning rates for the learner was initialized to 1e-3. The number of train steps for the learner (hyperparameter $k$) was set to 5. The meta-learner's learning rate was initialized to 3e-2 and set to linearly reduced to 3e-5 over 200 epochs. AdamW optimizer was used to train the meta-learner.

% \begin{table}[]
% \centering
% \caption{Results in 1-shot setting}
% \begin{tabular}{|c|c|c|}
% \hline
% \textbf{Setting}                                                                          & \multicolumn{1}{c|}{\textbf{Dice}} & \multicolumn{1}{l|}{\textbf{Parameter Count}} \\ \hline
% FSS-1000                                                                          & 73.47                     & 32.83M                               \\ \hline
% \begin{tabular}[c]{@{}c@{}}Ours,\\ CE + Dice loss,\\ threshold 0.5\end{tabular}  & 58.99                     & \multirow{4}{*}{\begin{tabular}[c]{@{}c@{}}9.56M\\ (For the learner. \\The meta learner has twice \\as many parameters)\end{tabular}}               \\ \cline{1-2}
% \begin{tabular}[c]{@{}c@{}}Ours,\\ CE + Dice loss,\\ threshold 0.35\end{tabular} & 63.08                     &                                      \\ \cline{1-2}
% \begin{tabular}[c]{@{}c@{}}Ours, \\ Cross Entropy,\\ threshold 0.5\end{tabular}  & 66.36                     &                                      \\ \cline{1-2}
% \begin{tabular}[c]{@{}c@{}}Ours, \\ Cross Entropy, \\ threshold 0.35\end{tabular} & 70.36                     &                                      \\ \hline
% \end{tabular}
% \label{tab:1}
% \end{table}
% \usepackage{multirow}

\subsection{Results}
% We observe that changing the threshold for setting sigmoid outputs to 1 has a significant impact on our model's performance with respect to the Dice metric. Specifically, we see that the best performance is reached at a threshold of around 0.35. It is also worth noting that the proposed model sees rather insignificant gain in Dice score on going from 1-shot to a 5-shot setting. 
The feature maps produced by each layer group in our model is visualized in Figure 2. An added advantage of not downsampling the feature maps, is that the shape of the target object is always clearly visible and the model's actions are easier to humanly interpret. The model steadily increases the separation between the object and the background as the feature maps progress through the layers. The blurry output of the degridding layers is due to the nature of upsampling performed by the sub-pixel convolution layer. 

\begin{figure}[t]
\begin{center}
   \includegraphics[width=\linewidth]{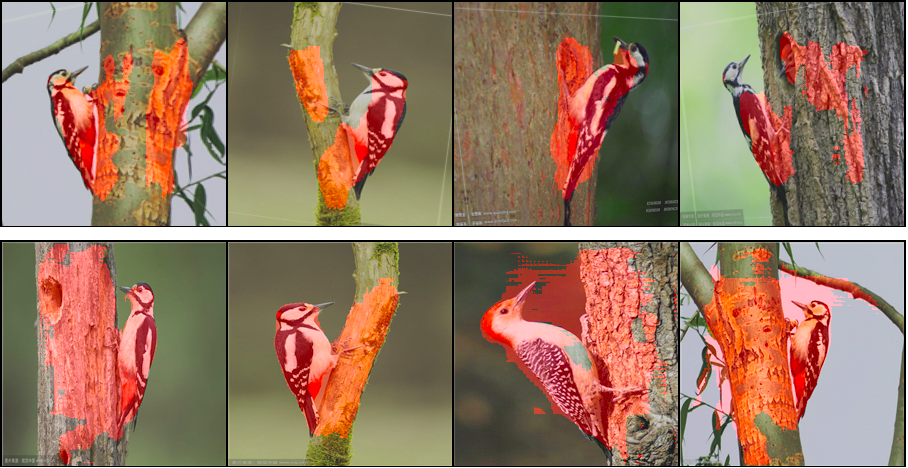}
\end{center}
   \caption{Visualization of a few validation images after 10 epochs of training. The top row is Meta-SGD model and bottom is Reptile.}
\label{fig:long}
\label{fig:onecol}
\end{figure}

\begin{figure*}[t]
  \centering \includegraphics[width=1\textwidth]{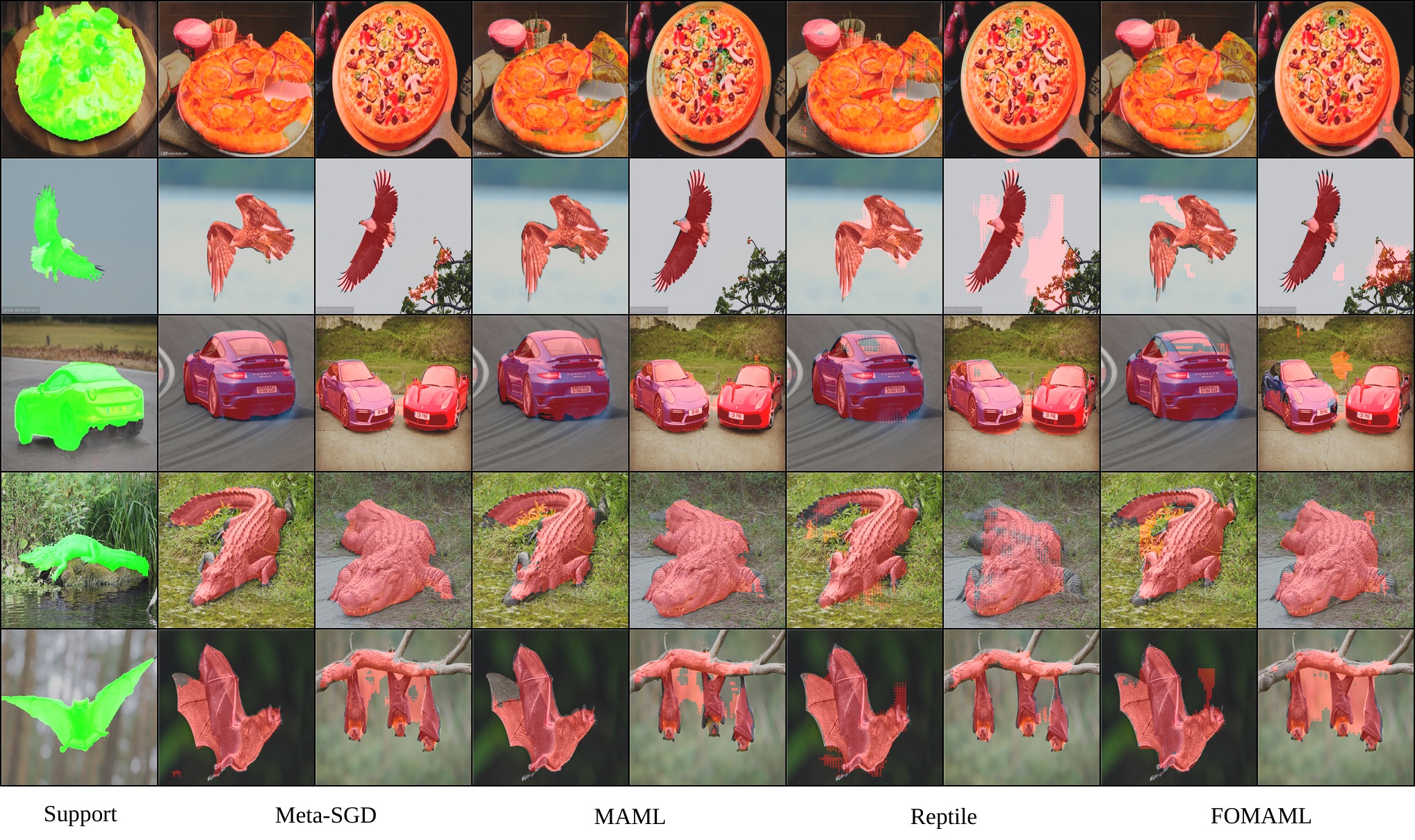}
  \caption{Results of 1-shot segmentation on some classes from the FSS-1000 test set. The images with green masks are the support images and ones with red masks are predictions on the corresponding query sets. The dataset classes from top to bottom are \textit{pizza}, \textit{eagle}, \textit{ferrari911}, \textit{american\_alligator}, and \textit{bat}.}
%   \caption{Results of 1-shot segmentation on some classes from the FSS-1000 test set. The images with green masks are the support images and ones with red masks are predictions on the corresponding query sets. The dataset classes from top to bottom are 'pizza', 'eagle', 'ferrari911', 'american\_alligator', and 'bat'}
\end{figure*}

The training for all 4 our meta-models were rather unstable and there were sudden drops and subsequent recovery in the validation IoU metric over the course of training. This phenomenon has been observed before for MAML in image classification. We observe that the non-thresholded mIoU is more stable than thresholded mIoU and hence a better validation metric to monitor. 

For image classification, Finn et al. have noted that FOMAML has a negligible loss in performance compared to MAML. Reptile too is able to reach similar accuracies as FOMAML and MAML. We observe that this is not the case for our chosen model architecture, where the second-order algorithms, MAML and Meta-SGD, outperformed FOMAML and Reptile. Although the second-order algorithms were approximately 50\% slower, they performed around 10\% better on the mIoU metric. As noted by Antoniou et al.\cite{maml++}, the performance of MAML is strongly dependent on model architecture. Changes to the model architecture may be able to bring the first-order algorithms closer to second-order ones. 

We observe that the Reptile model tends to produce final outputs with high pixel values while Meta-SGD tends to produce outputs with lower pixel values. In Figure 3, we visualize a few validation images from both Meta-SGD and Reptile after 10 epochs of training. All masks are generated by setting threshold to $0.5$. Meta-SGD tends to generate smaller masks that partially cover the target objects while Reptile generates larger masks that cover more than target objects. This effect persists even after the end of training. The mean IoU for Meta-SGD goes up on lowering the threshold from 0.5 while it goes down for Reptile. In Table 1 we show the results for a lower threshold of 0.35(arbitrarily selected) along with the standard threshold of 0.5. MAML outputs tend to be in-between Reptile and Meta-SGD.

The final results for our approach along with the existing benchmark on the FSS-1000 dataset has been shown in Table 1. We present our results with 95\% confidence intervals. Our best performing model was MetaDRN trained with Meta-SGD. Its expected mIoU value on the FSS-1000 test was 76.63\%, which is 3.16\% higher than the benchmark of 73.47\%.

\section{Conclusion}
In this paper we propose a lightweight CNN architecture for 1-shot image segmentation. The model is based on dilated residual networks\cite{drn} and uses dilated convolutions in place of downsampling. Unlike the original DRN model we use a sub-pixel convolution in place of deconvolution for upsampling the final feature maps.

We train our model on the FSS-1000 using 4 meta-learning algorithms that work well for image classification and get results comparable to the existing benchmark. We also find that for our architecture, 2nd-order meta-learning algorithms perform better than 1st-order algorithms, in the same settings. 
% In this paper we propose a lightweight CNN architecture for 1-shot image segmentation. The model makes use of dilated convolutions in place of downsampling to increase accuracy, and reduce the number of operations needed. To further speed up the model, a sub-pixel convolution layer\cite{PixelShuffle} is used for upsampling instead of using a deconvolution layer. 

% Our model trains a meta-learner using Meta-SGD\cite{msgd} algorithm. The meta-learner learns to initialize a learner's weights and learning rates for a new task. On the FSS-1000 dataset\cite{fss}, our model achieves a dice score within 4\% of the benchmark, while having a 70\% lower parameter count.

\bibliographystyle{IEEEtran}
\bibliography{bibliography.bib}
% No cites here
% \nocite{reza}
% \nocite{hendryx}
\nocite{relational_nets}
\nocite{nocite}
\nocite{nocite2}
\nocite{nocite3}
\nocite{nocite4}
\nocite{nocite5}
\nocite{wandb}
\nocite{FastAI}
\end{document}